\documentclass[11pt]{article}

\usepackage[preprint]{acl}

\usepackage{times}
\usepackage{latexsym}

\usepackage[T1]{fontenc}

\usepackage[utf8]{inputenc}

\usepackage{microtype}

\usepackage{inconsolata}

\usepackage{multirow}
\usepackage[table]{xcolor}
\usepackage{longtable}
\usepackage{booktabs}
\usepackage{listings}
\usepackage{wrapfig}
\usepackage{rotating}

\usepackage{amsmath,amssymb,amsthm}

\usepackage{pgf}

\usepackage{graphicx}
\usepackage[utf8]{inputenc} 
\usepackage[T1]{fontenc}    
\usepackage{hyperref}       
\usepackage{url}            
\usepackage{booktabs}       
\usepackage{amsfonts}       
\usepackage{nicefrac}       
\usepackage{microtype}      
\usepackage{xcolor}         
\usepackage{amsmath}
\usepackage{amssymb}
\usepackage{graphicx}
\usepackage{enumitem}
\usepackage{makecell}

\usepackage[most,skins,theorems]{tcolorbox}
\tcbset{
  aibox/.style={
    width=\linewidth,
    colback=blue!6!white,
    colframe=black,
    colbacktitle=black,
    enhanced,
    center,
    attach boxed title to top left={yshift=-0.1in,xshift=0.15in},
    boxed title style={boxrule=0pt,colframe=white,},
    minipage boxed title=0.825\linewidth, %
  },
  promptstyle/.style={enhanced,colback=white,colframe=black,colbacktitle=gray!20,
    coltitle=black,rounded corners,sharp corners=north,boxrule=0.5pt,
    drop shadow=black!50!white,attach boxed title to top left={xshift=-2mm,yshift=-2mm},
    boxed title style={rounded corners,size=small,colback=gray!20}},
  replystyleg/.style={enhanced,colback=green!15,colframe=black,colbacktitle=green!30,
    coltitle=black,boxrule=0.5pt,drop shadow=black!50!white,rounded corners,
    sharp corners=north,attach boxed title to top right={xshift=-2mm,yshift=-2mm},
    boxed title style={rounded corners,size=small,colback=green!40}},
  replystyler/.style={enhanced,colback=red!15,colframe=black,colbacktitle=red!40,
    coltitle=black,boxrule=0.5pt,drop shadow=black!50!white,rounded corners,
    sharp corners=north,attach boxed title to top right={xshift=-2mm,yshift=-2mm},
    boxed title style={rounded corners,size=small,colback=red!40}}
}
\newtcolorbox{AIbox}[2][]{aibox,title=#2,#1}
\newtcolorbox{promptbox}[1][]{promptstyle,title=Prompt,#1}

\title{Simulating Environments with Reasoning Models for Agent Training}

\author{
\textbf{Yuetai Li}\textsuperscript{$\clubsuit$}\thanks{Work done during an internship at Microsoft.} \;\;\;  
\textbf{Huseyin A Inan}\textsuperscript{$\spadesuit$} \;\;\; 
\textbf{Xiang Yue}\textsuperscript{$\diamondsuit$} \;\;\; 
\textbf{Wei-Ning Chen}\textsuperscript{$\spadesuit$} \;\;\; 
\textbf{Lukas Wutschitz}\textsuperscript{$\spadesuit$} \;\;\; \\
\textbf{Janardhan Kulkarni}\textsuperscript{$\spadesuit$} \;\;\; 
\textbf{Radha Poovendran}\textsuperscript{$\clubsuit$}
\textbf{Robert Sim}\textsuperscript{$\spadesuit$} \;\;\; 
\textbf{Saravan Rajmohan}\textsuperscript{$\spadesuit$} \;\;\; \\
  \textsuperscript{$\clubsuit$}University of Washington \; 
  \textsuperscript{$\spadesuit$}Microsoft \;
  \textsuperscript{$\diamondsuit$}Carnegie Mellon University \; \\
  \texttt{\{yuetaili,rp3\}@uw.edu},\\
  \texttt{\{huinan,weiningchen,luwutsch,jakul,rsim,saravar\}@microsoft.com},\\
  \texttt{xyue2@andrew.cmu.edu}
  ~ \\
   \textbf{Huggingface}: \url{https://huggingface.co/Simia-Agent} \\ 
   \textbf{Github}: \url{https://github.com/microsoft/Simia-Agent-Training} \\
   ~ \\
   ~ \\
}

\begin{document}

\maketitle

\begin{abstract}

LLM agents excel in compact environments requiring deep reasoning but remain brittle when operating in broader, more complex contexts that demand robustness across diverse tools and schemas. Building bespoke environments for training is heavy, brittle, and limits progress. In this paper, we demonstrate that LLMs can simulate realistic environment feedback without access to actual testbed data or APIs. Inspired by this capability, we propose two frameworks: \textbf{Simia-SFT}, a pipeline that synthesizes SFT data by amplifying small seed sets into diverse trajectories in an environment-agnostic manner, and \textbf{Simia-RL}, a framework that enables RL training without real environment implementations through LLM-simulated feedback. Fine-tuning open models yields consistent improvements across multiple benchmarks, surpassing GPT-4o and approaching o4-mini on $\tau^2$-Bench. Together, Simia-SFT and Simia-RL enable scalable agent training without environment engineering, replacing heavy and brittle implementations with flexible LLM-based simulation.

\end{abstract}

\section{Introduction}
\label{sec:introduction}

Large language model (LLM) agents are becoming increasingly capable and autonomous.
Advances in reasoning mechanisms have enabled them to perform multi-step planning, tool use, and problem solving, to the point of surpassing human experts in narrow but cognitively demanding domains such as mathematics competitions (e.g. the International Mathematical Olympiad \citep{LuongLockhart2025}) or programming contests (e.g. the ICPC \citep{LinCheng2025}).
We refer to these as \textit{complex-task/simple-environment} settings: the core challenge lies in solving a difficult reasoning problem, while the environment itself is well-specified, compact, and easily set up.

In contrast, LLM agents continue to struggle with tasks that are individually simple but situated in broad, messy, or dynamic contexts.
Examples include operating vending machines, office workflows, or household tasks \citep{backlund2025vendingbenchbenchmarklongtermcoherence}.
We call these \textit{simple-task/complex-environment} settings.
The reasoning required for each step may be minimal, yet success depends on robustness across a wide distribution of environments, tools, and edge cases.

\begin{figure}
    \vspace{2em}
    \centering
    \includegraphics[width=\linewidth]{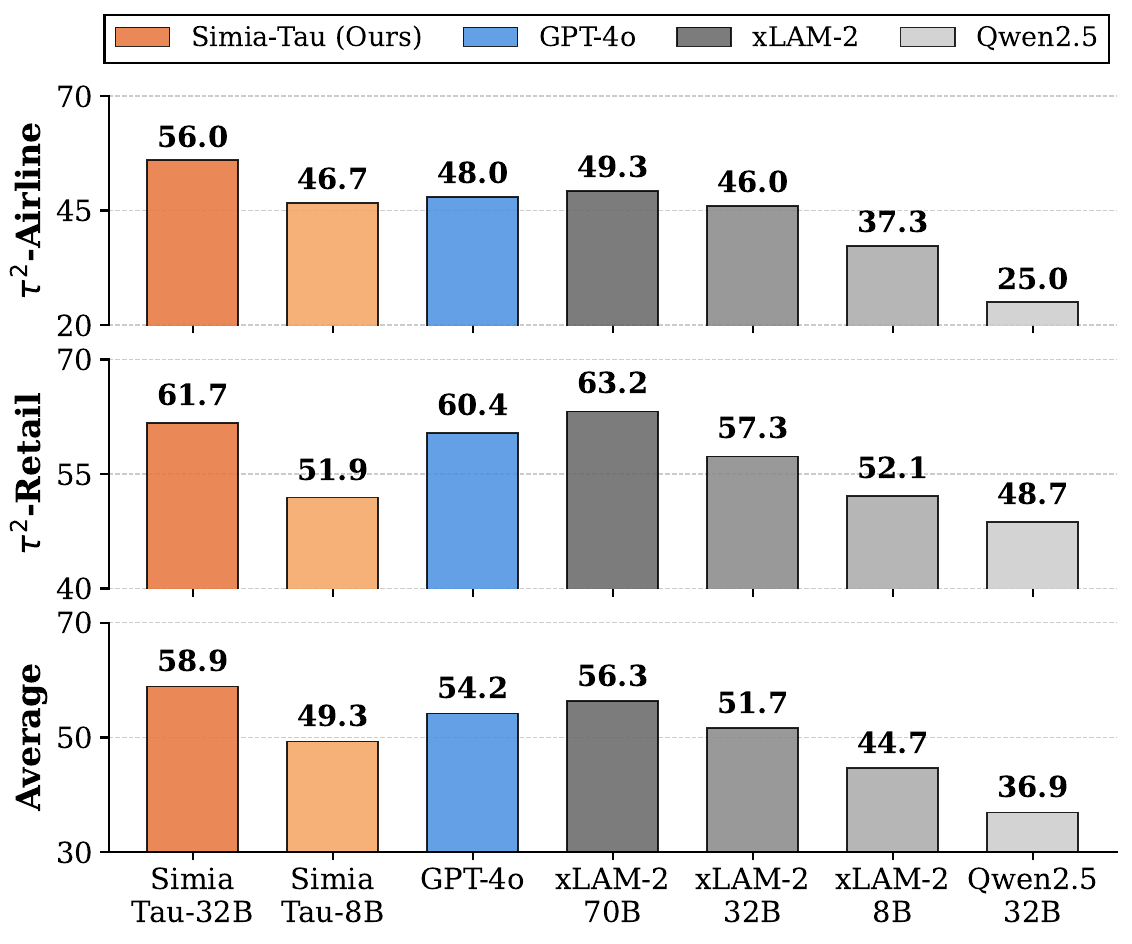}
    \caption{Performance of models fine-tuned on our synthetic simulated trajectories without real environment implementations. Our 32B model (based on Qwen2.5-32B-Instruct) surpasses GPT-4o and xLAM-2-70B model and our 8B model (based on Qwen3-8B) outperforms Qwen2.5-32B-Instruct on $\tau^2$-Airline and Retail.}
    \label{fig:teaser}
\end{figure}

We argue that training agents on diverse trajectories spanning a broad range of environments is essential for tackling simple-task/complex-environment challenges \citep{kimiteam2025kimik2openagentic}. Yet building such data at scale is non-trivial. 
Synthetic data generation offers a promising alternative for scalable trajectory creation \citep{patil2023gorillalargelanguagemodel, tang2023toolalpacageneralizedtoollearning, qin2023toolllmfacilitatinglargelanguage, zeng2023agenttuningenablinggeneralizedagent}. However, existing synthesis pipelines are heavily environment-dependent. Traditional approaches construct agent trajectories by implementing environments with explicit tool interfaces and APIs \citep{patil2023gorillalargelanguagemodel, zeng2023agenttuningenablinggeneralizedagent}. While this yields realistic interactions, it tightly couples data generation to environment-specific details, limiting transferability and scalability. Each new environment demands substantial engineering to set up and define tool specifications, making broad coverage impractical.

In this paper, we show that LLMs can act as environment simulators, generating coherent state transitions and tool interactions without access to real environments. This reveals LLMs' capability as environment simulators, exploiting their world modeling abilities to generate coherent environment dynamics, state transitions, and tool interactions without access to the real agent environments.

Building on this observation, we propose: \textbf{Simia-SFT}, a pipeline that synthesizes SFT data by amplifying small seed sets into diverse trajectories in an environment-agnostic manner, and \textbf{Simia-RL}, a framework that enables RL training without real environment implementations by generating LLM simulated feedback. Together, Simia-SFT and Simia-RL enable scalable agent training across diverse tasks without the burden of agentic environment engineering, replacing heavy and brittle environment implementations with flexible LLM-based simulation.

Fine-tuning open models such as \texttt{Qwen3-8B} and \texttt{Qwen2.5-32B-Instruct} on simulated trajectories delivers substantial gains across multiple benchmarks~\citep{wang2024officebench, barres2025tau, liu2023agentbench}. Figure~\ref{fig:teaser} shows that
on $\tau^2$-Bench, our 32B model surpasses GPT-4o and xLAM-2-70B \cite{prabhakar2025apigen} and our 8B model outperforms Qwen2.5-32B-Instruct on Airline and Retail.

In summary, this paper makes the following contributions:
\begin{itemize}[leftmargin=*]
\item \textbf{Simia-SFT Pipeline}. We propose an environment-agnostic trajectory synthesis pipeline that generates complete agent trajectories from small seed sets without executing real environments.
\item \textbf{Simia-SFT Dataset}. The open-source tool-agent training dataset covers multi-turn dialogues and tool use, contributing agentic training and improving agent performance. 
\item \textbf{Simia-RL Framework}. We propose RL on LLM-simulated environments that circumvents agent environment engineering.
\item \textbf{Empirical gains.} Across diverse benchmarks, open models fine-tuned on simulated trajectories and RL on simulated environment yield consistent improvements, surpassing closed-source counterparts on the different benchmarks.
\end{itemize}

These results suggest a practical recipe for agent training: replace environment-specific code with LLM simulators. 
This reframes environment engineering as an amortized prompt-and-schema design question, enabling broader progress in agentic LLM training.

\begin{figure}
    \centering
    \includegraphics[width=\linewidth]{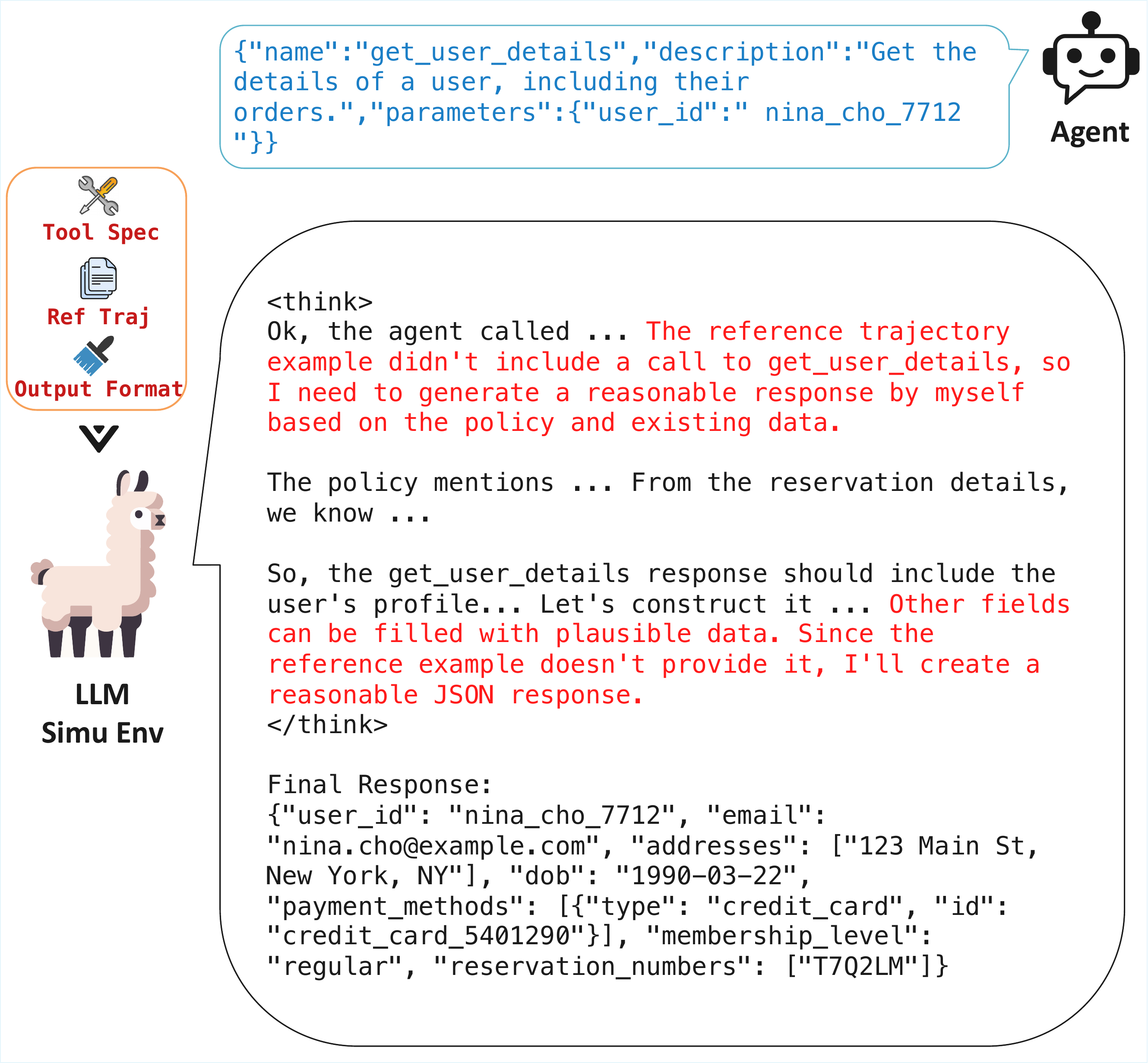}
    \caption{LLM can reason to simulate plausible environment feedback without requiring access to all actual testbed data or system information.}
    \label{fig:llm_simulation}
\end{figure}

\begin{figure*}[htbp]
    \centering
    \includegraphics[width=\textwidth]{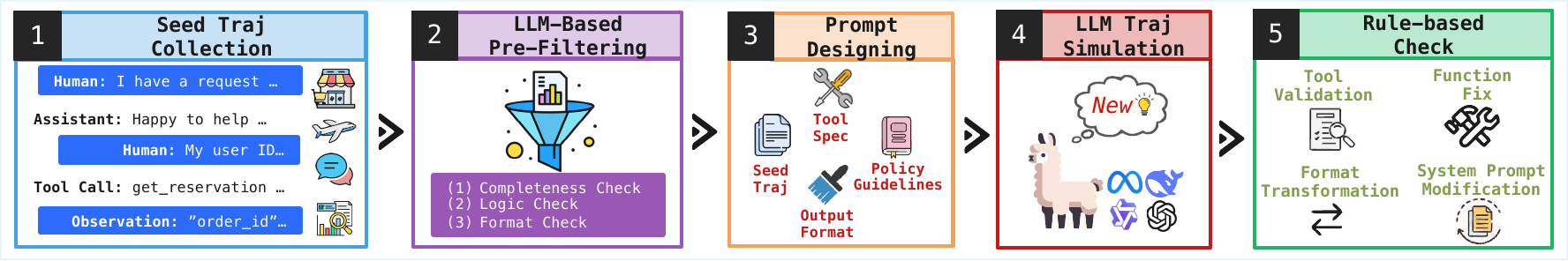}
    \caption{Simia-SFT pipeline to synthesize agent trajectory data without real environment executions. The diagram shows the flow from seed trajectory through pre-filtering, prompt design,  LLM simulation and final sanity check.}
    \label{fig:framework_overview}
\end{figure*}
\section{LLM Simulator for Agent Training}

In this section, we demonstrate that LLMs can reason to simulate agent environments even without the actual testbed data. By leveraging this capability, we propose: (1) \textbf{Simia-SFT}, a pipeline that synthesizes agent SFT trajectories in an environment-agnostic manner, and (2) \textbf{Simia-RL}, a framework that enables RL on simulated environment.

\subsection{LLMs Can Reason to Simulate Agent Environments}

Figure~\ref{fig:llm_simulation} shows one example that LLMs can simulate realistic environment feedback when provided with interaction history, tool usage specifications, reference trajectory and environment response formats, even without access to actual testbed data. This demonstrates LLMs' capability as environment simulators, exploiting their world modeling abilities to generate coherent environment dynamics, state transitions, and tool interactions without access to the real agent environments. 

Building on this observation, we develop LLM trajectory simulator for SFT in Section \ref{sec:SFT_methodology} and LLM environment simulator for RL in Section \ref{LLM Environment Simulator for RL}.

\subsection{Simia-SFT: SFT with LLM simulated Traj}
\label{sec:SFT_methodology}

Traditional approaches construct agent SFT trajectories by implementing real environments with explicit tool interfaces and APIs. While this yields realistic interactions, it tightly couples the trajectory generation process to environment-specific details, limiting transferability and scalability. 

We address the problem of generating synthetic trajectories in a manner that is independent of agent environments. We proposed an \textbf{End-to-end Agent Trajectory Synthesis} pipeline called \textbf{Simia-SFT}, where complete trajectories are simulated directly from seed data without requiring real environment implementations.

\subsubsection{End-to-end Agent Trajectory Synthesis Pipeline}

Figure~\ref{fig:framework_overview} illustrates our pipeline for agent SFT data synthesis. Given a set of seed trajectory, our framework generates agent trajectories through a four-stage process: (1) \textbf{LLM-based Pre-filtering} to validate seed quality, (2) \textbf{Prompt Design} to anchor generation to valid action spaces, (3) \textbf{LLM Trajectory Simulation} to synthesize diverse multi-round interactions, and (4) \textbf{Rule-based Checks} to ensure structural correctness. Each component is described in detail below.

\paragraph{LLM-Based Pre-Filtering.}
Before trajectory synthesis, we validate seed trajectories through LLM-based evaluation across three dimensions: (i) \textit{Completeness Check} verifies that tasks are successfully completed, with all necessary components including task descriptions, reasoning steps, tool invocations, and environment responses; (ii) \textit{Logic Check} assesses the consistency of reasoning chains and action sequences; (iii) \textit{Format Check} validates structural adherence, including proper role round and correct JSON formatting. Seeds passing all checks form a curated set for trajectory simulation, while those with inconsistencies are discarded.

\paragraph{Prompt Design.}
To ensure generated trajectories remain valid without accessing to real environments, we incorporate the formal specification of the environment's tool and API interfaces directly into the prompt. This anchors generation to the correct action space. The prompt contains: (i) available tool specifications, (ii) policy rules and constraints, (iii) expected input and output format, and (iv) one seed reference trajectory as an exemplar. 
The complete prompt templates are provided in Appendix~\ref{appendix:prompt}.

\paragraph{LLM Trajectory Simulation.}
The LLM simulator is prompted to synthesize novel agent trajectories inspired by seed examples. In a single generation call, the model simulates a complete agent trajectory, producing alternating user queries, agent reasoning, tool invocations, and simulated environment observations until task completion.
We employ temperature adjustment and multiple generation passes to promote diversity, amplifying small seed sets into diverse trajectories that vary in task formulation, reasoning strategies, and action sequences. This enables scalable training data creation without real environment execution.

\paragraph{Rule-Based Post-Process.}
We apply rule-based post-processing to ensure structural validity. First, we repair malformed function arguments by fixing common JSON syntax errors (e.g., unescaped quotes, missing brackets). Second, we validate that tool calls reference available tools and filter out invalid trajectories. We also normalize tool calling by using Hermes XML style (e.g., \texttt{<tool\_call>}) or JSON objects, discarding trajectories that cannot be reliably parsed. Finally, we modify system prompts to match the target deployment environment. This post-processing ensures training-ready data with the necessary structural fidelity.


Our approach differs from traditional distillation by expanding task distributions and enforcing structural fidelity absent from raw synthesizer outputs, while remaining bounded by the synthesizer's semantic correctness.

\begin{figure}
    \centering
    \includegraphics[width=0.7\linewidth]{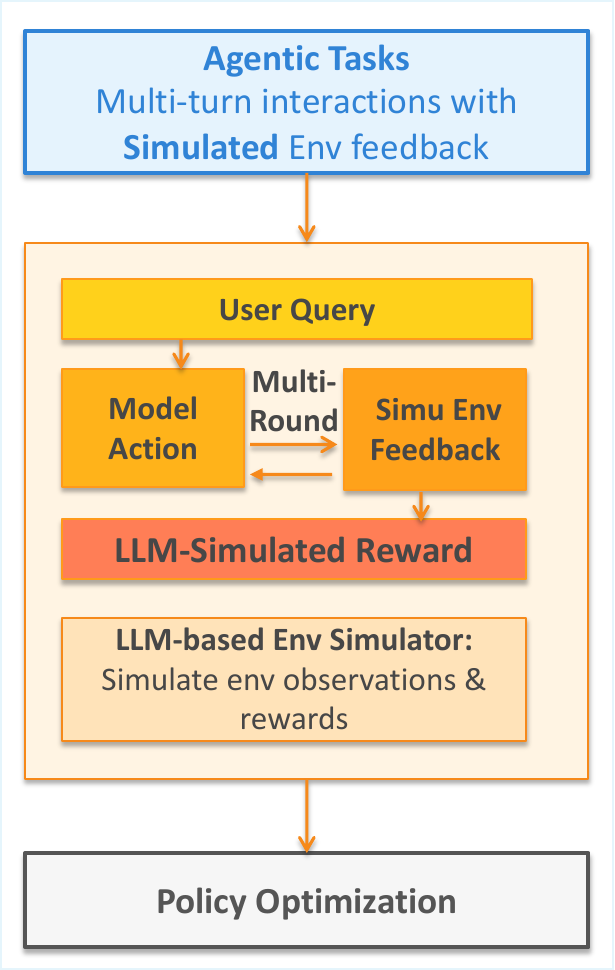}
    \caption{Simia-RL framework, which enables RL through multi-turn interactions within simulated environments. An LLM-based simulator provides both environment feedback and reward signals to support iterative policy optimization.}
    \label{fig:rl_framework_overview}
\end{figure}

\subsection{Simia-RL: RL on LLM Simulated Env}
\label{LLM Environment Simulator for RL}

RL for agent training requires models to interact with environments, computing rewards from these interactions to guide policy optimization. However, existing agent RL are typically environment-specific, heavily dependent on custom implementations where each scenario demands dedicated environment setup, tool interfaces, and reward functions. This necessitates deploying and maintaining separate setups for each agent environment, preventing agent RL across diverse scenarios within a unified framework and limiting scalability.

Inspired by LLMs' capability to simulate environment feedback, we propose \textbf{Simia-RL}, an agentic RL framework for RL on LLM-simulated environments that circumvents per-task environment engineering. Figure~\ref{fig:rl_framework_overview} illustrates our RL framework with multi-turn interactions. We incorporate tool usage specifications, environment feedback formats, reference trajectory samples, and interaction history into the LLM prompt. The simulator comprises two key components: (1) \textbf{Environment Feedback Simulation} processes agent actions to produce simulated tool outputs and error messages; (2) \textbf{Reward Computation} assesses trajectory completion and assigns reward based on predefined success criteria and task objectives.

\definecolor{ourmodel}{RGB}{0, 0, 0}  
\definecolor{lightblue}{RGB}{220, 240, 255}  
\definecolor{gray}{RGB}{128, 128, 128}
\definecolor{lightgray}{RGB}{240, 240, 240}

\begin{table}
\centering
\caption{
Performance on $\tau^2$-Bench (Airline, Retail) across proprietary models, open baselines, and our fine-tuned models. Our SFT models (\textbf{Simia-Tau}) consistently outperform baselines and narrow the gap to or even surpass much larger proprietary models. Sequential SFT followed by RL on simulated environments (\textbf{Simia-Tau-RL}) yields slightly additional gains over SFT alone.
}
\label{tab:taubench}
\resizebox{\linewidth}{!}{
\begin{tabular}{l|ccc}
\toprule
\textbf{Model} & \textbf{Airline} & \textbf{Retail} & \textbf{Average} \\
\midrule
\multicolumn{4}{c}{\textit{GPT Models}} \\
\midrule
GPT-5 & \textbf{58.0} & \textbf{77.2} & \textbf{67.6} \\
o4-mini & \underline{57.0} & \underline{69.3} & \underline{63.2} \\
GPT-4.1 & 53.0 & 65.2 & 59.1 \\
GPT-4o & 48.0 & 60.4 & 54.2 \\
GPT-4 & 36.3 & 54.4 & 45.4 \\
\midrule
\multicolumn{4}{c}{\textit{$\geq$ 32B Models}} \\
\midrule
\rowcolor{lightblue}
\textbf{\textcolor{ourmodel}{Simia-Tau (Qwen2.5-32B)}} & \textbf{\textcolor{ourmodel}{56.0}} & \textbf{\textcolor{ourmodel}{61.7}} & \textbf{\textcolor{ourmodel}{58.9}} \\
xLAM-2-70B & \underline{49.3} & \textbf{63.2} & \underline{56.3} \\
xLAM-2-32B & 46.0 & \underline{57.3} & 51.7 \\
Qwen2.5-32B-Instruct & 25.0 & 48.7 & 36.9 \\
\midrule
\multicolumn{4}{c}{\textit{7B/8B Models}} \\
\midrule
\rowcolor{lightblue}
\textbf{\textcolor{ourmodel}{Simia-Tau-RL (Qwen3-8B)}} & \textbf{\textcolor{ourmodel}{49.0}} & \textbf{\textcolor{ourmodel}{52.9}} & \textbf{\textcolor{ourmodel}{51.0}} \\
\rowcolor{lightblue}
\textbf{\textcolor{ourmodel}{Simia-Tau (Qwen3-8B)}} & \underline{\textcolor{ourmodel}{46.7}} & 51.9 & \underline{\textcolor{ourmodel}{49.3}} \\
xLAM-2-8B & 37.3 & \underline{52.1} & \underline{44.7} \\
\addlinespace[3pt]
\rowcolor{lightblue}
\textbf{\textcolor{ourmodel}{Simia-Tau-RL (Qwen2.5-7B)}} & \textcolor{ourmodel}{41.5} & \textcolor{ourmodel}{44.0} & \textcolor{ourmodel}{42.8} \\
\rowcolor{lightblue}
\textbf{\textcolor{ourmodel}{Simia-Tau (Qwen2.5-Coder-7B)}} & \textcolor{ourmodel}{41.3} & \textcolor{ourmodel}{43.4} & \textcolor{ourmodel}{42.4} \\
\addlinespace[3pt]
\rowcolor{lightblue}
\textbf{\textcolor{ourmodel}{Simia-Tau (Qwen2.5-7B)}} & \textcolor{ourmodel}{39.3} & \textcolor{ourmodel}{40.9} & \textcolor{ourmodel}{40.1} \\
\addlinespace[3pt]
\rowcolor{lightblue}
\textbf{\textcolor{ourmodel}{Simia-Tau (Llama-3.1-8B)}} & \textcolor{ourmodel}{36.0} & \textcolor{ourmodel}{36.3} & \textcolor{ourmodel}{36.2} \\
Qwen3-8B-APIGen-MT-5k & 22.7 & 48.7 & 35.7 \\
Qwen2.5-7B-Instruct-APIGen-MT-5k & 22.0 & 38.7 & 30.4 \\
Qwen3-8B & 16.0 & 31.6 & 23.8 \\
Qwen2.5-7B-Instruct & 18.0 & 11.4 & 14.7 \\
Qwen2.5-Coder-7B-Instruct & 24.0 & 5.3 & 14.7 \\
Llama-3.1-8B-Instruct & 22.0 & 5.3 & 13.7 \\
\midrule
\multicolumn{4}{c}{\textit{3B Models}} \\
\midrule
\rowcolor{lightblue}
\textbf{Simia-Tau (Llama-3.2-3B)} & \textbf{\textcolor{ourmodel}{36.0}} & \textbf{\textcolor{ourmodel}{32.0}} & \textbf{\textcolor{ourmodel}{34.0}} \\
\rowcolor{lightblue}
\textbf{\textcolor{ourmodel}{Simia-Tau (Qwen2.5-3B)}} & \underline{\textcolor{ourmodel}{32.3}} & \underline{\textcolor{ourmodel}{31.1}} & \underline{\textcolor{ourmodel}{31.7}} \\
xLAM-2-3B & 19.3 & 18.7 & 19.0 \\
Llama-3.2-3B-Instruct & 30.0 & 6.1 & 18.1 \\
Qwen2.5-3B-Instruct & 12.7 & 6.2 & 9.5 \\
\midrule
\multicolumn{4}{c}{\textit{1.5B Models}} \\
\midrule
\rowcolor{lightblue}
\textbf{\textcolor{ourmodel}{Simia-Tau (Qwen2.5-1.5B)}} & \underline{24.0} & \textbf{\textcolor{ourmodel}{17.6}} & \textbf{\textcolor{ourmodel}{20.8}} \\
xLAM-2-1.5B & \textbf{30.7} & \underline{6.7} & \underline{18.7} \\
Qwen2.5-1.5B-Instruct & 22.0 & 5.3 & 13.7 \\
\bottomrule
\end{tabular}
}
\vspace{-1em}
\end{table}

\section{Experiments}
\label{sec: experiments}
\subsection{Setup}

\paragraph{SFT Setup.}
We synthesize data from three seed datasets: (i) APIGen-MT \citep{prabhakar2025apigen}, which contains $\sim$1.5k training trajectories in distribution with $\tau$-bench Airline and $\sim$3.5k in distribution with $\tau$-bench Retail~\citep{yao2024tau}); from this seed we generate 90k trajectories. (ii) AgentTuning~\citep{zeng2023agenttuning}, which provides 195 samples for Operating System~\citep{liu2023agentbench}, 351 for WebShop~\citep{yao2022webshop}, and 122 for Mind2Web~\citep{deng2023mind2web}; we expand these into 15k trajectories spanning the three domains. (iii) OfficeBench~\citep{wang2024officebench}, where we take 76 1-app tasks with o4-mini trajectories as the seed and synthesize 30k samples targeting multi-app settings. Trajectory simulation is performed using GPT-5 and o4-mini as synthesizers with temperature 1.0.

We fine-tune the synthesized trajectories on the Qwen2.5/3 and Llama~3.1/3.2 model families across multiple sizes. Each trajectory is segmented at assistant turns, and training is performed to predict only assistant tokens while masking prompts and other messages. Fine-tuning is conducted with LLaMA-Factory~\citep{zheng2024llamafactory} under a full-parameter setting, with hyperparameter details provided in Appendix~\ref{SFT Setup}.

\begin{figure}
  \centering
  \includegraphics[width=0.9\linewidth]{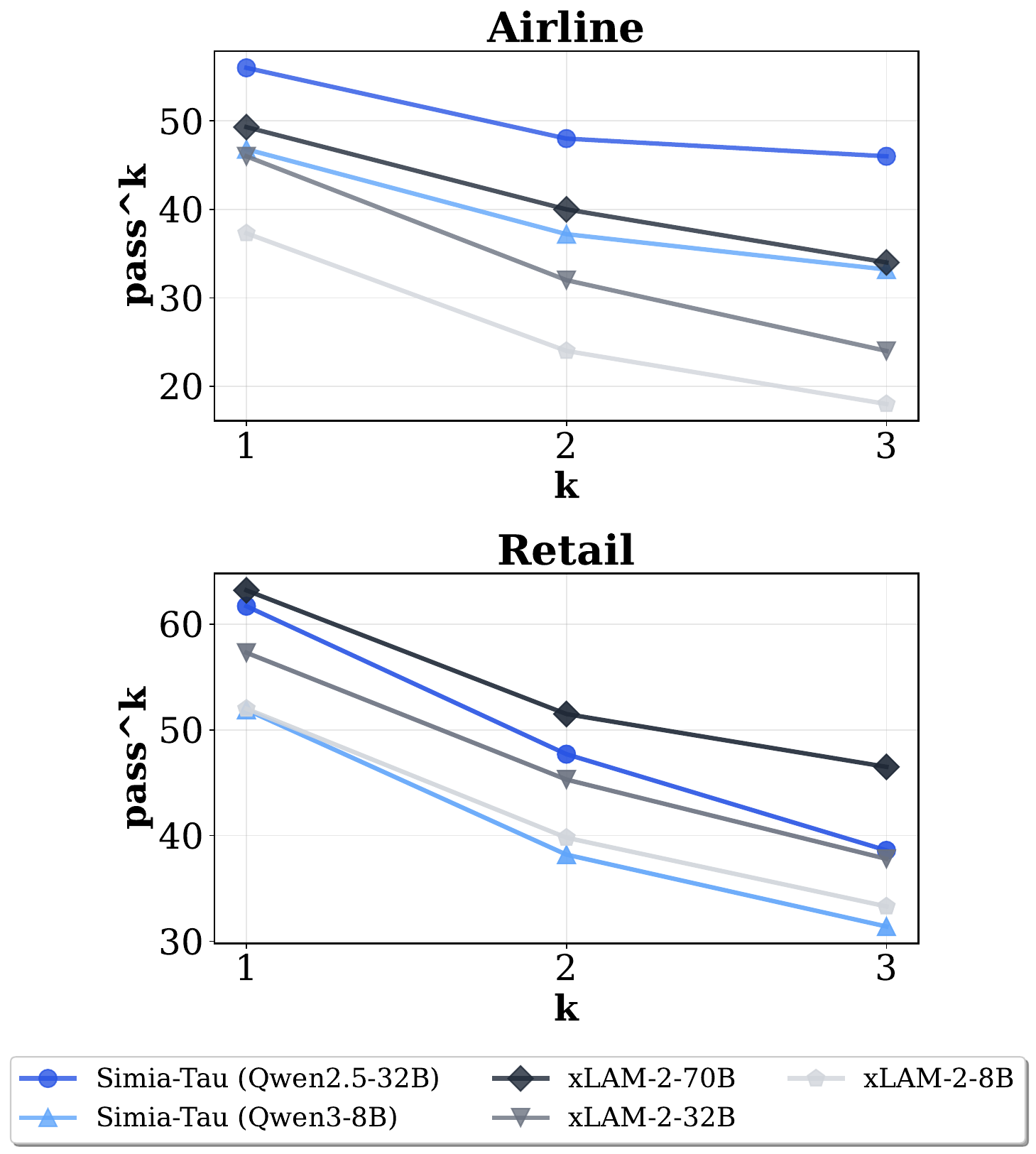}

\caption{ \texttt{Pass\^{}k} performance comparison on the $\tau^2$-Bench across Airline and Retail domains for Simia-Tau and xLAM-2 models, with $k$ values of 1, 2, and 3. \texttt{Pass\^{}k} requires that each task should be successful for all the k retries, highlighting the robustness.}

  \label{fig:tau_bench_pass_power_k_comparison}
\end{figure}

\paragraph{Benchmarks.}
To assess the generality of our approach, we evaluate on three complementary agentic suites that span distinct domains and interaction types: (1) $\tau^2$-Bench (Airline, Retail)~\citep{barres2025tau}, a realistic tool-use benchmark emphasizing multi-turn API invocation, error recovery, and state tracking. Compared with $\tau$-Bench~\citep{yao2024tau}, $\tau^2$-Bench~\citep{barres2025tau} corrects potential erroneous tasks and enhances system implementation. Following \citep{barres2025tau}, we use GPT-4.1 as the user simulator (temperature set to 0) to generate user behaviors.
(2) OfficeBench~\citep{wang2024officebench} (2-apps, 3-apps), which evaluates cross-application workflows, compositional tool use, and coordination across office utilities.
(3) AgentBench~\citep{liu2023agentbench} (Operating System, WebShop, Mind2Web), which spans software manipulation, e-commerce goal completion, and open-web browsing.
Together, these suites comprehensively stress multi-step reasoning and tool-use competency across distinct domains.

\paragraph{Baselines.}
For reference, we report results from proprietary models including GPT-4, GPT-4o, GPT-4.1, o4-mini, and GPT-5. For open-source baselines, we fine-tune Qwen2.5-7B-Instruct and Qwen3-8B on the seed datasets of each suite; APIGen-MT (5k), AgentTuning (668), and OfficeBench 1-app tasks with o4-mini trajectories (76) using the same training recipe as our method. We further include the xLAM-2 model family~\citep{prabhakar2025apigen} and AgentLM family~\citep{zeng2023agenttuningenablinggeneralizedagent}, spanning model sizes from 1B to 70B.

\paragraph{RL Setup.} We implement RL experiments using RAGEN \citep{wang2025ragen} built on VeRL \citep{sheng2024hybridflow}. 
We conduct GRPO~\citep{shao2024deepseekmath} training followed by SFT under two configurations: (1) training on OfficeBench 1-app tasks and evaluating on OfficeBench 2-apps and 3-apps tasks, and (2) training on APIGen-MT-5k and evaluating on $\tau^2$-Bench airline and retail. We run GRPO training for 64 steps in both configurations. Detailed RL training hyperparameters are provided in Appendix~\ref{Grpo Setup}.

We utilize o4-mini to simulate the environment, which interacts with the model in each round, and computes the final reward. o4-mini is configured with a temperature of 1.0 and a maximum context length of 60,000 tokens. The complete tool usage specification, environment feedback format, the interaction history, and one reference sample trajectory are converted into text and incorporated into the o4-mini's prompt. 
Upon task completion or reaching the maximum number of interaction rounds, o4-mini will assess whether the task was successfully completed based on the trajectory. We assign a reward of 1 for successful task completion and 0 for failure. Detailed prompt are shown in Appendix \ref{appendix:prompt_rl}.

\subsection{Results}

\paragraph{$\tau^2$-bench.}
Table \ref{tab:taubench} shows that our Simia-Tau models demonstrate substantial performance gains across the Airline and Retail domains of the $\tau^2$-Bench. Notably, Simia-Tau (Qwen2.5-32B) achieves an average score of 58.9 (56.0 on Airline, 61.7 on Retail), outperforming GPT-4o by 4.7 points while trailing GPT-4.1 by only 0.2 points and o4-mini by 4.3 points. It also surpasses the baseline xLAM-2-32B (average 51.7) by 7.2 points and even exceeds xLAM-2-70B (average 56.3) by 2.6 points.

\definecolor{ourmodel}{RGB}{0, 0, 0}  
\definecolor{lightblue}{RGB}{220, 240, 255}  
\definecolor{gray}{RGB}{128, 128, 128}
\definecolor{lightgray}{RGB}{240, 240, 240}
\begin{table}
    \centering
    \caption{
        Performance on OfficeBench~\citep{wang2024officebench} across proprietary models, open baselines, and models fine-tuned on simulated trajectories. SFT followed by RL on simulated environments (\textbf{Simia-OB-RL}) also yields slight gains over SFT-only models (\textbf{Simia-OB}).
    }
    \label{tab:officebench_comparison}
    \resizebox{1\linewidth}{!}{
        \begin{tabular}{l|ccc}
            \toprule
            \textbf{Model} & \textbf{2-apps} & \textbf{3-apps} & \textbf{Average} \\
            \midrule
            \multicolumn{4}{c}{\textit{GPT Models}} \\
            \midrule
            GPT-5 & \underline{84.3} & \textbf{76.4} & \textbf{80.4} \\
            o4-mini & \textbf{86.3} & \underline{70.9} & \underline{78.6} \\
            GPT-4.1 & 78.4 & 63.6 & 71 \\
            GPT-4o & 74.5 & 50.9 & 62.7 \\
            GPT-4 & 50.6 & 11.6 & 31.1 \\
            \midrule
            \multicolumn{4}{c}{\textit{7B/8B Models}} \\
            \midrule
            \rowcolor{lightblue}
            \textbf{\textcolor{ourmodel}{Simia-OB-RL (Qwen2.5-7B)}} & \underline{64.7} & \textbf{34.5} & \underline{49.6} \\
            \rowcolor{lightblue}
            \textbf{\textcolor{ourmodel}{Simia-OB (Qwen3-8B)}} & 58.8 & \textbf{29.1} & \textbf{44.0} \\
            \addlinespace[3pt]
            \rowcolor{lightblue}
            \textbf{\textcolor{ourmodel}{Simia-OB (Qwen2.5-7B)}} & 57.8 & 27.3 & \underline{42.6} \\
            \rowcolor{lightblue}
            \textbf{\textcolor{ourmodel}{Simia-OB (Qwen2.5-Coder-7B)}} & \underline{64.7} & 14.6 & 39.7 \\
            \rowcolor{lightblue}
            \textbf{\textcolor{ourmodel}{Simia-OB (Llama-3.1-8B)}} & \textbf{64.9} & 12.7 & 33.8 \\
            Qwen3-8B-1apps-seed & 39.2 & 20 & 29.6 \\
            Qwen2.5-7B-Instruct-1apps-seed & 35.3 & 16.4 & 25.9 \\
            Qwen2.5-Coder-7B-Instruct & 31.4 & 10.7 & 21.1 \\
            Qwen2.5-7B-Instruct & 27.5 & 10.9 & 19.2 \\
            Llama-3.1-8B-Instruct & 7.8 & 3.6 & 5.7 \\
            Qwen3-$8B$ & 3.9 & 0 & 2.0 \\
            \midrule
            \multicolumn{4}{c}{\textit{3B Models}} \\
            \midrule
            \rowcolor{lightblue}
            \textbf{\textcolor{ourmodel}{Simia-OB (Qwen2.5-3B)}} & \textbf{43.1} & \textbf{9.1} & \textbf{26.1} \\
            \rowcolor{lightblue}
            \textbf{\textcolor{ourmodel}{Simia-OB (Llama-3.2-3B)}} & \textbf{43.1} & \underline{7.3} & \underline{25.2} \\
            Qwen2.5-3B-Instruct & \underline{15.7} & 1.8 & 8.8 \\
            Llama-3.2-3B-Instruct & 7.6 & 0 & 3.8 \\
            \midrule
            \multicolumn{4}{c}{\textit{1.5B Models}} \\
            \midrule
            \rowcolor{lightblue}
            \textbf{\textcolor{ourmodel}{Simia-OB (Qwen2.5-1.5B)}} & \textbf{33.3} & \textbf{5.5} & \textbf{19.4} \\
            Qwen2.5-1.5B-Instruct & \underline{5.9} & \underline{0} & \underline{3.0} \\
            \bottomrule
        \end{tabular}
    }
\end{table}

For 7B/8B-scale models, Simia-Tau (Qwen3-8B) attains an average score of 49.3 (46.7 on Airline, 51.9 on Retail), outperforming GPT-4 by 3.9 points and xLAM-2-8B (average 44.7) by 4.6 points. Furthermore, our Qwen3-8B substantially outperform models fine-tuned on APIGen-MT-5k, exceeding Qwen3-8B-APIGen-MT-5k (average 35.7) by 13.6 points. Additionally, after SFT and RL, Simia-tau-RL (Qwen-8B) achieves 49.0 and 52.9 on Ailine and Retail.

Figure \ref{fig:tau_bench_pass_power_k_comparison} illustrates the \verb|Pass^k|~\citep{barres2025tau} performance of Simia-Tau and xLAM-2 models, evaluating the effect of varying $k$ values (1, 2, and 3) on task success rates. \verb|Pass^k|~\citep{barres2025tau} requires that each task should be successful for all the k retries to measure the robustness. Our Simia-Tau (Qwen2.5-32B) consistently achieves higher pass rates, with scores of 56.0, 48.0, and 46.0 for Airline, and 61.7, 47.7, and 38.6 for Retail, outperforming xLAM-2-70B Airline (49.3, 40.0, 34.0) and slightly less for Retail (63.2, 51.5, 46.5) as $k$ increases.

\definecolor{ourmodel}{RGB}{0, 0, 0}  
\definecolor{lightblue}{RGB}{220, 240, 255}  
\definecolor{gray}{RGB}{128, 128, 128}
\definecolor{lightgray}{RGB}{240, 240, 240}
\begin{table}
    \centering
    \caption{
        Performance on Webshop, Mind2Web and Operating System~\citep{liu2023agentbench} across proprietary models, open baselines, and models fine-tuned on simulated trajectories.
        *GPT-4 results are reported from \citep{liu2023agentbench} and GPT-4o results are reported from leaderboard of SuperBench \citep{superbench_modeldetail}. 
    }
    \label{tab:agentbench_comparison}
    \resizebox{\linewidth}{!}{
        \begin{tabular}{l|ccc|c}
            \toprule
            \textbf{Model} & \textbf{Mind2Web} & \textbf{OS} & \textbf{Webshop} & \textbf{Average} \\
            \midrule
            \multicolumn{5}{c}{\textit{GPT Models}} \\
            \midrule
            GPT-4* & \underline{29} & \textbf{42.4} & \textbf{61.1} & \textbf{44.2} \\
            GPT-4o* & \textbf{32} & \underline{33.3} & \underline{48.9} & \underline{38.1} \\
            \midrule
            \multicolumn{5}{c}{\textit{$\geq$7B Models}} \\
            \midrule
            \rowcolor{lightblue}
            \textbf{\textcolor{ourmodel}{Simia-AB (Qwen3-8B)}} & \underline{25} & \textbf{34.5} & 67.3 & \textbf{42.6} \\
            \rowcolor{lightblue}
            \textbf{\textcolor{ourmodel}{Simia-AB (Qwen2.5-Coder-7B)}} & \textbf{29} & \underline{33} & 66 & \underline{42.6} \\
            \rowcolor{lightblue}
            \textbf{\textcolor{ourmodel}{Simia-AB (Qwen2.5-7B)}} & 23 & 32.6 & \underline{69.2} & 41.6 \\
            Qwen3-8B-seed & 22 & 31.3 & 68.6 & 40.6 \\
            Qwen2.5-7B-Instruct-seed & 12 & 29.1 & 68.2 & 36.4 \\
            AgentLM-70B & 13.5 & 21.5 & 64.9 & 33.3 \\
            Qwen3-8B & 21 & 27.8 & 49.5 & 32.8 \\
            AgentLM-13B & 8.4 & 18.1 & \textbf{70.8} & 32.4 \\
            Qwen2.5-7B-Instruct & 7 & 27.8 & 58.8 & 31.2 \\
            Llama-3.1-8B-Instruct & 18 & 19.4 & 54.9 & 30.8 \\
            Qwen2.5-Coder-7B-Instruct & 17 & 22 & 53 & 30.5 \\
            AgentLM-7B & 6.4 & 17.4 & 63.6 & 29.1 \\
            \midrule
            \multicolumn{5}{c}{\textit{3B Models}} \\
            \midrule
            \rowcolor{lightblue}
            \textbf{\textcolor{ourmodel}{Simia-AB (Llama-3.2-3B)}} & \textbf{26} & \underline{28} & \textbf{69} & \textbf{40.9} \\
            \rowcolor{lightblue}
            \textbf{\textcolor{ourmodel}{Simia-AB (Qwen2.5-3B)}} & \underline{23} & \textbf{31} & \underline{65} & \underline{39.9} \\
            Qwen2.5-3B-Instruct & 16 & 24 & 49 & 29.5 \\
            Llama-3.2-3B-Instruct & 15 & 17 & 49 & 27.0 \\
            \midrule
            \multicolumn{5}{c}{\textit{1.5B Models}} \\
            \midrule
            \rowcolor{lightblue}
            \textbf{\textcolor{ourmodel}{Simia-AB (Qwen2.5-1.5B)}} & \underline{17} & \textbf{15} & \textbf{54} & \textbf{28.5} \\
            Qwen2.5-1.5B-Instruct & \textbf{20} & \underline{13} & \underline{40} & \underline{24.0} \\
            \bottomrule
        \end{tabular}
    }
\end{table}

\paragraph{OfficeBench.}
Table~\ref{tab:officebench_comparison} shows that our Simia-OB models demonstrate substantial improvements in office automation tasks on the OfficeBench, which evaluates language agents' ability to handle complex workflows involving multiple applications such as Word, Excel, calendars, and emails. Notably, Simia-OB (Qwen3-8B) achieves an average score of 44.0 (58.8 on 2-apps, 29.1 on 3-apps), outperforming GPT-4 (average 31.1) by 12.9 points. Other variants, such as Simia-OB (Qwen2.5-7B) with 42.6 average and Simia-OB (Qwen2.5-Coder-7B) with 39.7, further highlight the efficacy of our approach in bridging the gap between small models and proprietary giants. For the Qwen3-8B model, the $\text{thinking}$ behavior leads to excessively long CoT sequences and repetitive patterns, which results in performance scores approaching zero. 

Compared with the model fine-tuned on the seed data with real environments, Simia-OB (Qwen3-8B)  outperforms baselines Qwen3-8B-1apps-seed (average 29.6) by 14.4 points. Similarly, Simia-OB (Qwen2.5-7B) exceeds Qwen2.5-7B-Instruct-1apps-seed (average 25.9) by 16.7 points.

\begin{figure*}
    \centering
    \includegraphics[width=\textwidth]{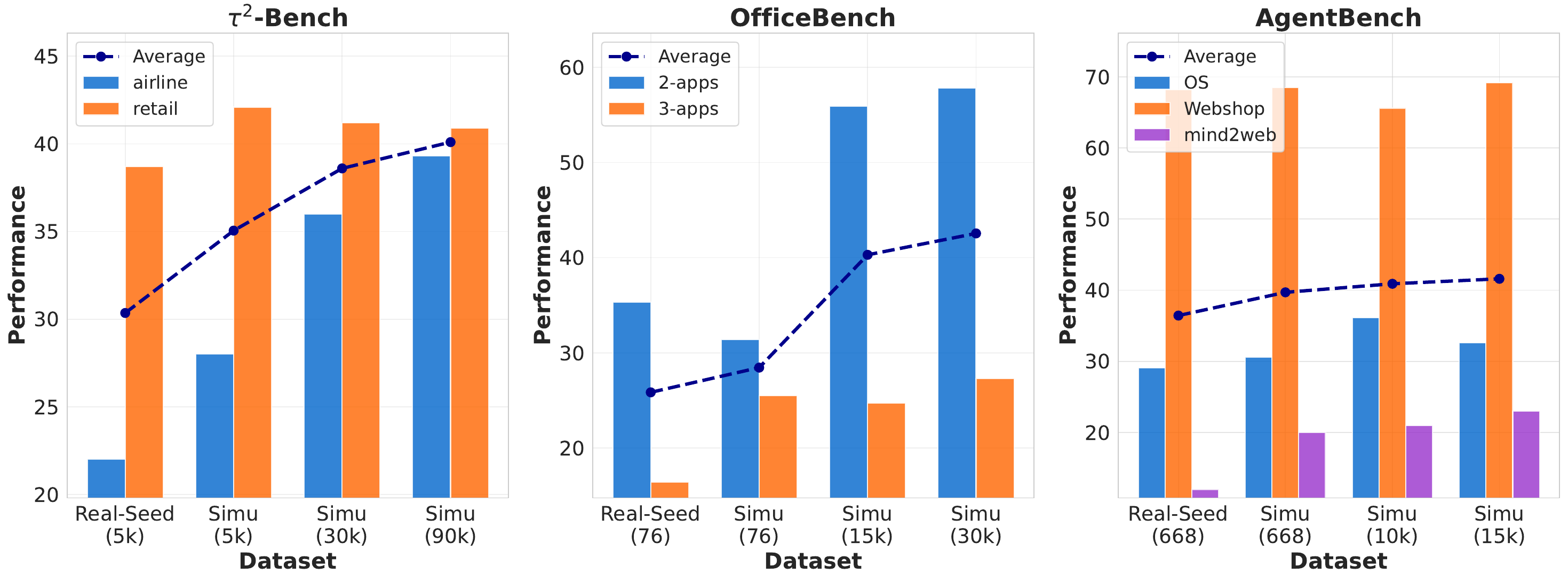}
    \caption{Ablation study on dataset generated by real environment and simulated environment. When the dataset size is identical, we show that simulated trajectories achieve performance comparable to real-environment-based trajectories on OfficeBench and AgentBench, and even better on $\tau^2$-bench. As dataset size scales, simulated trajectories significantly improve model performance. This highlights the potential of simulated environments to support larger, more diverse datasets, addressing the limitations of real-world seed data constrained by collection efforts. }
    \label{fig:ablation_on_datasize}
\end{figure*}

\paragraph{AgentBench.} Our models exhibit competitive performance across Mind2Web, Operating System (OS), and Webshop, as presented in Table~\ref{tab:agentbench_comparison}. Our Simia-AB (Qwen3-8B) achieves an average score of 42.6, with a standout 34.5 on OS and 67.3 on Webshop, closely matching GPT-4’s 44.2 average and surpassing GPT-4o’s 38.1. The Simia-AB (Qwen2.5-Coder-7B) model ties with Qwen3-8B at 42.6 average, excelling with 29 on Mind2Web and 66 on Webshop, though it lags slightly on OS (33). The Qwen2.5-7B model leads with a Webshop score of 69.2, outperforming GPT-4 (61.1) and GPT-4o (48.9), with an overall average of 41.6. Both of our fine-tuned Qwen3-8B and Qwen2.5-7B-instruct models are better than their variants finetuned on the seed dataset generated from real environment.

\begin{figure*}
  \centering
  \includegraphics[width=\textwidth]{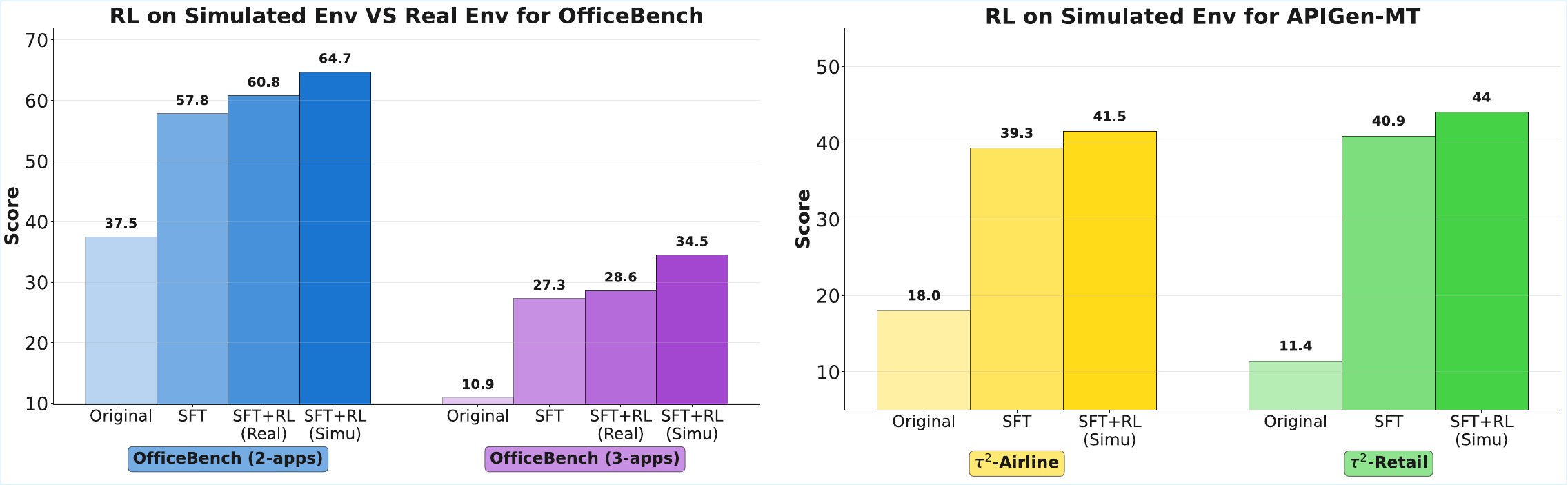}
  \caption{RL on simulated environment followed by SFT for Qwen2.5-7B-Instruct model.}
  \label{fig:sft_rl_officebench}
\end{figure*}

\paragraph{Trajectory by Real Env V.S. Simulated Env.} Figure \ref{fig:ablation_on_datasize} shows the performance of Qwen2.5-7B-Instruct fine-tuned on trajectories generated from real environment (seed dataset) and simulated environments (our synthesized dataset). When trained on datasets of identical size (e.g., for $\tau^2$-bench we compare the model fine-tuned on 5k seed data and 5k synthesized data), we find that synthesized trajectories achieve performance comparable to real-environment-based trajectories on OfficeBench and AgentBench, even better on $\tau^2$-bench. As dataset size expands, synthesized trajectories significantly outperform seed-based ones, demonstrating the scalability advantage of synthesized data with simulated environments. This highlights the potential of simulated environments to support larger, more diverse datasets, addressing the limitations of real-world seed data constrained by collection efforts.

\begin{figure*}
    \centering
    \includegraphics[width=0.9\textwidth]{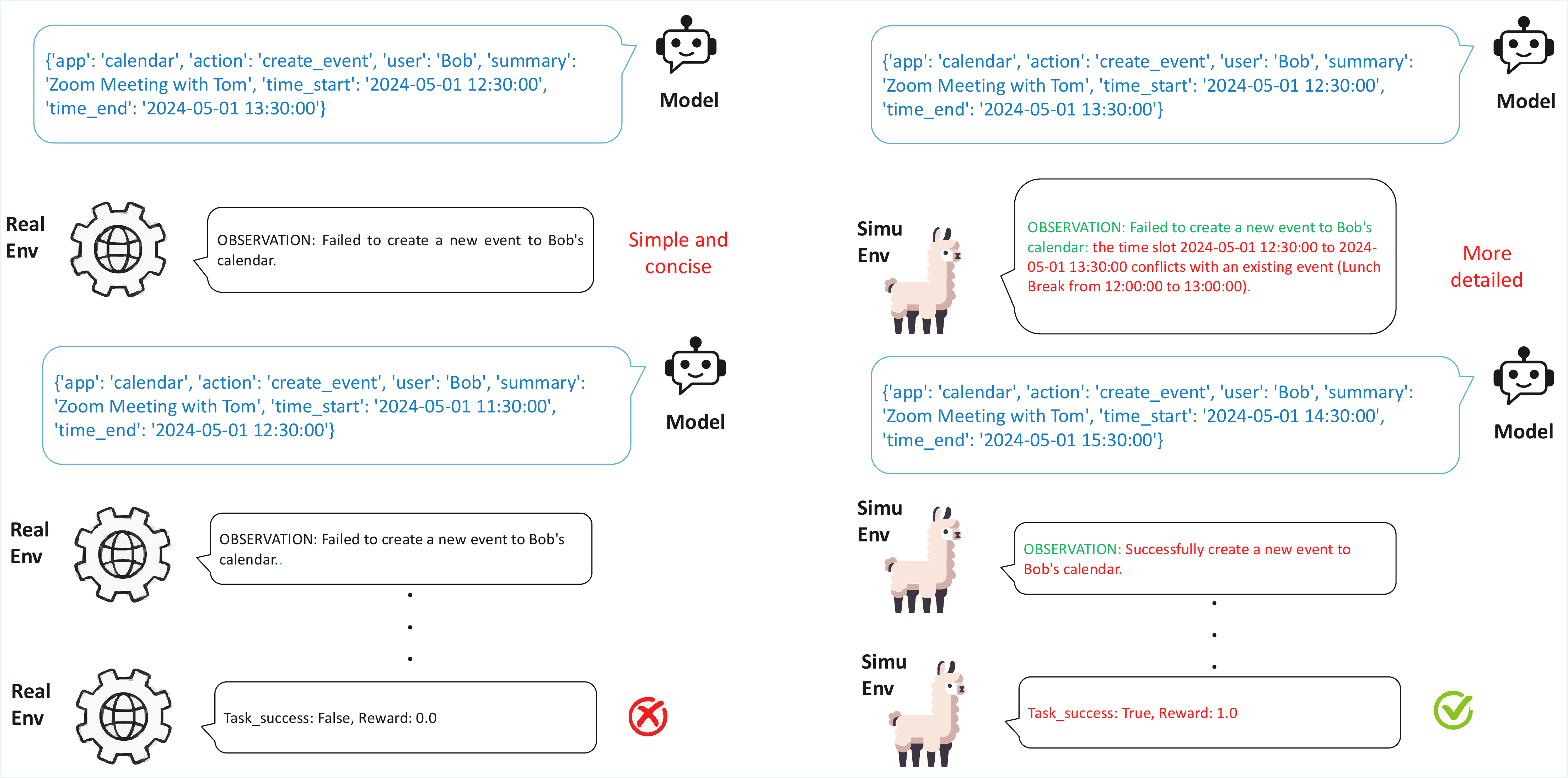}
    \caption{Case study comparing RL on real and simulated environments.
In the real setting (left), the system displays only the fixed failure messages inherently provided by the original environment \cite{wang2024officebench}. In contrast, the simulated environment (right) generates more flexible and detailed feedback, explaining conflicts (e.g., overlapping with a lunch break) and enabling the model to adjust and eventually get the reward.}
    \label{fig:rl_case_study}
\end{figure*}

\subsection{RL Results}
\paragraph{Results of RL on Simulated Env.} Figure \ref{fig:sft_rl_officebench} demonstrates the results of RL on simulated environments for Qwen2.5-7B-Instruct. For OfficeBench, We observe that the simulated environment (64.7, 34.5) outperforms the real environment (60.8 28.6) on OfficeBench 2-apps and 3-apps, yielding total improvements of 6.9 and 7.2 points over the model after SFT. On $\tau^2$-Bench, SFT followed by RL yields slight additional gains. Note that $\tau^2$-Bench real environment requires simulated users for dynamic agent interaction, we do not directly compare simulated and real environments for $\tau^2$-Bench.

In the real setting (left), the system displays only the fixed and concise failure messages inherently provided by the original environment.

\paragraph{Case Study of RL on Simulated Env for OfficeBench.} 
We present in Figure \ref{fig:rl_case_study} a case study highlighting how RL surprisingly benefits more from the simulated environment than from the real one. In the real setting (left), the system displays only the fixed and concise failure messages inherently provided by the original environment \cite{wang2024officebench}. By contrast, the simulated environment (right) offers richer and more adaptive feedback, e.g., pointing out conflicts such as overlapping with a lunch break, which helps the model adjust its behavior and ultimately achieve the reward.

\subsection{More Ablations}
\paragraph{Ablation on Different Trajectory Simulators.}
We compare performances of different trajectory simulators in Appendix \ref{More Results of Synthetic Data}.

\paragraph{Training on Combined Datasets.}
We train Qwen3-8B on combined three datasets jointly. See more details in Appendix~\ref{More Results of Synthetic Data}.

\section{Conclusion}
In this work, we observed that LLMs can reason to simulate agent environments even without the actual testbed data. By leveraging this capability, we propose Simia-SFT, an LLM trajectory simulator framework to synthesize agentic SFT data. By combining LLm-based pre-filtering, LLM simulation and rule-based sanity check, our pipeline produces high-quality trajectories across heterogeneous agent environments without requiring laborious environment engineering. We further proposed Simia-RL, an LLM environment simulator framework to enable RL without real agent environment implementations. Our experiments show that smaller open models surpass much larger proprietary models. Together, these results establish environment simulation as a practical pathway for advancing LLM agent training, opening opportunities for scalable research and deployment across diverse real-world tasks.

\subsubsection*{Limitations}
While our pipeline targets broad, environment-agnostic trajectory synthesis, our current study still has several limitations. Experiments focus on a finite set of agent domains such as airline, retail, web navigation, and office scenarios. Extending to more domains with different tool schemas is important. Additionally, our data generation and RL environments rely on LLM simulators. Although prompts constrain the action space by anchoring formal tool specifications, the simulation may still introduce distributional bias that differ from real deployments.

\subsubsection*{Ethical Considerations}
This work investigates synthetic data and simulated environments for agentic training.  
It does not involve human subjects, user studies, or the collection of personally identifiable information. 
The dataset and benchmarks used in our experiments are publicly available. We do not introduce or endorse any applications that could cause harm or be misused.
This paper does not present any ethical concerns.

\bibliography{acl2026}


\clearpage
\appendix

\section{Related Work}
\label{sec:related_work}
\paragraph{Tool-using LLMs.} Despite strong reasoning and generation abilities, LLMs fall short on tasks requiring real-time access, accurate computation, or environment interaction, motivating the development of \emph{LLM-based agents} that leverage external tools. 
Representative examples include LLMs that use a web-browsing environment for question answering~\citep{nakano2022webgptbrowser}, a Python interpreter for arithmetic and symbolic reasoning~\citep{gao2023palprogramaidedlanguagemodels}, and information retrieval for grounding in dialogue~\citep{thoppilan2022lamdalanguagemodelsdialog}. 
The range of tools explored in the literature is far broader, and we refer readers to the comprehensive survey in~\citet{Qu_2025} for full coverage.

\paragraph{Synthetic agentic datasets.} While tool-using LLMs demonstrate the promise of agents, scaling their development requires large volumes of training data. To this end, recent efforts have introduced synthetic agentic datasets that emulate queries, tools, and environment feedback at scale \citep{patil2023gorillalargelanguagemodel,tang2023toolalpacageneralizedtoollearning,qin2023toolllmfacilitatinglargelanguage,zeng2023agenttuningenablinggeneralizedagent,xu2025toucan,li-etal-2023-api}. Gorilla~\citep{patil2023gorillalargelanguagemodel} uses a self-instruct approach to generate instruction–API pairs. ToolAlpaca~\citep{tang2023toolalpacageneralizedtoollearning} leverages a multi-agent simulation environment to build a diverse tool-use corpus, while ToolLLM~\citep{qin2023toolllmfacilitatinglargelanguage} employs ChatGPT to synthesize instruction–solution pairs involving real-world RESTful APIs. AgentTuning~\citep{zeng2023agenttuningenablinggeneralizedagent} adopts GPT-4 as an agent to generate trajectories across six task domains. API-Bank~\citep{li2023apibankcomprehensivebenchmarktoolaugmented} uses a multi-agent pipeline to generate domains, simulate APIs, construct queries, and validate responses. APIGen~\citep{liu2024apigenautomatedpipelinegenerating}, built on ToolBench~\citep{qin2023toolllmfacilitatinglargelanguage}, applies multi-stage verification with diverse prompt templates to ensure accuracy and coverage. ToolBridge~\citep{jin2024toolbridgeopensourcedatasetequip} curates a dataset of Python-based tool invocations via selection, conversion, and filtering of public corpora. BUTTON~\citep{chen2025facilitatingmultiturnfunctioncalling} generates multi-turn function-calling data with GPT-4o, combining bottom-up task evolution with top-down decomposition of complex tasks. Finally, ToolACE~\citep{liu2025toolacewinningpointsllm} introduces a pipeline for synthesizing diverse, complex, and accurate function-calling data through iterative tool evolution, complexity-guided dialog synthesis, and dual-layer verification.

Existing synthetic datasets have made significant progress in enabling tool-augmented LLMs, introducing diverse APIs, generation pipelines, and verification strategies. Our work takes a complementary direction: we simulate complete end-to-end trajectories that integrate reasoning steps, tool use, and environment feedback. This benchmark-agnostic design allows synthetic data to be extended across diverse domains and tasks, providing a flexible way to augment existing datasets with richer, trajectory-level supervision. In this sense, our method does not replace prior efforts but instead offers a means of extending their coverage and contributes agentic training.

\section{Experiment Setup}
\label{Detailed Experiment Setup}

\subsection{Supervised Fine-tuning}
\label{SFT Setup}

Our model SFT is conducted using LLaMA-Factory \citep{zheng2024llamafactory}, on a server with eight NVIDIA A100-SXM4-80GB GPUs. We follow \cite{prabhakar2025apigen} for the training parameters.
Table \ref{tab: training-hyperparameters} lists hyper-parameters for full parameter supervised fine-tuning.

\begin{table}[!h]
\small
\centering
\caption{Hyper-parameters used for full parameter supervised fine-tuning.}
\resizebox{0.8\columnwidth}{!}{
\begin{tabular}{ll}
\toprule
\textbf{Hyper-parameter} & \textbf{Value} \\ \midrule
Learning Rate & $5 \times 10^{-6}$ \\
Number of Epochs & $2$ \\
Number of Devices & $8$ \\
Per-device Batch Size & $2$ \\
Optimizer & \texttt{Adamw} \\
Learning Rate Scheduler & \texttt{cosine} \\
Max Sequence Length  & $16384$ \\ \bottomrule
\end{tabular}
}
\label{tab: training-hyperparameters}
\end{table}

\subsection{GRPO}
\label{Grpo Setup}

We use the following hyper-parameters detailed in Table \ref{tab: rl hyperparameters} for RL training on the simulated environment. We perform experiments on eight A100 GPUs. The model is trained using RAGEN \citep{ragen}and VeRL \citep{sheng2024hybridflow}. The RL training is configured with a rollout of 16, 64 training steps, a maximum context length of 12,000 tokens, and up to 40 interaction rounds per task, with a temperature setting of 0.7. 

\begin{table}[htbp]
\small
\centering
\caption{Hyper-parameters for RL training.}
\begin{tabular}{ll}
\toprule
\textbf{Hyper-parameter} & \textbf{Value} \\ \midrule
Learning Rate & $1 \times 10^{-6}$ \\
Number of Steps & $64$ \\
Number of Devices & $8$ \\
Temperature & 0.7 \\
Top P & 1.0 \\
PPO Mini Batch Size & 32 \\
Max Response Length & $16384$ \\
KL Coefficient & $0.001$ \\
Rollout Engine & $\textsc{vllm (v0.8.2)}$ \\
Optimizer & \texttt{Adamw} \\
Learning Rate Scheduler & \texttt{cosine} \\
Warmup Ratio & $0.1$ \\
Agent Max Turn & 25 \\
Agent Max Action Per Turn & 1 \\
Clip Ratio High & 0.28 \\

\bottomrule
\end{tabular}
\label{tab: rl hyperparameters}
\end{table}

\section{Full Prompt for Synthetic Trajectory Generation}
\label{appendix:prompt}

See figure \ref{fig:Prompt to synthesize APIGen-MT} for synthesize APIGen-MT and Agenttuning, and see figure \ref{fig:Prompt to synthesize OfficeBench} for OfficeBench.

\begin{figure*}[htbp]
    \centering
\begin{tcolorbox}[title=Prompt to synthesize APIGen-MT and Agenttunig]
\begin{tcblisting}{
  title=,
  listing only,
  breakable,
  enhanced,
  colback=white,
  colframe=black!15,
  boxrule=0.3pt,
  listing options={
    basicstyle=\ttfamily\scriptsize, % ← 比 footnotesize 更小
    breaklines=true,
    columns=fullflexible
  }
}
You are an AI assistant that generates multi-turn conversation data for agent training. Your task is to create new agent trajectories based on existing examples.

## Example Trajectory:
{sample_text}

## Available Tools:
{available_tools}

CRITICAL FORMAT PRESERVATION REQUIREMENTS - ABSOLUTE COMPLIANCE:
1. **STRICTLY PRESERVE ORIGINAL FORMAT**: You MUST maintain the EXACT format structure from the example trajectory (EXCEPTION: function_call turns may include <think> tags when reasoning is needed)
2. **NO SYSTEM PROMPT GENERATION**: Do NOT generate any SYSTEM messages - follow the system prompt from the original example and that will be preserved separately
3. **TOOL CONSTRAINT ADHERENCE**: You MUST STRICTLY use ONLY the tools listed in the "Available Tools" section above. DO NOT use any tools outside this specified allowed tool set. This is MANDATORY.
4. **FORMAT CONSISTENCY**: Maintain identical conversation structure, role naming conventions, and response patterns as shown in the example (EXCEPTION: function_call turns may include <think> tags when reasoning is added)
5. **TURN COUNT MATCHING**: Generate approximately the SAME NUMBER of conversation turns as the example trajectory - the generated conversation should have a comparable length and depth to the sample data

## FUNCTION_CALL TURN REQUIREMENTS:
1. **REASONING IN THINK TAGS**: When making function calls, add brief reasoning (1-3 sentences) inside `<think> </think>` tags ONLY in FUNCTION_CALL turns after you output 'FUNCTION_CALL:'
2. **SELECTIVE REASONING**: Not every function call needs reasoning. Only include it when it helps explain the complex decision-making process
3. **STRICT TURN CONSTRAINT**: Reasoning in `<think> </think>` tags should ONLY appear in FUNCTION_CALL turns, NEVER in HUMAN, GPT, OBSERRVATION or additional turns
4. **FORMAT REQUIREMENT**: If reasoning is included, the FUNCTION_CALL turn are allowed to add the thinking sentences instead of only JSON format. You should follow this format:
'''
FUNCTION_CALL:
<think>
Brief reasoning about why this function call is needed (1-3 sentences). Ended with: I will call the function <function_name>.
</think>
{{"name": "function_name", "arguments": {{...}}}}
'''

## ABSOLUTE PROHIBITIONS:
- DO NOT use ANY tools that are not explicitly listed in the "Available Tools" section above
- DO NOT change the conversation format structure (human/gpt roles, value formatting, etc.) - EXCEPTION: function_call turns may include <think> tags when reasoning is needed
- DO NOT violate any fixed formatting elements, tool specifications, or requirements in system instructions from the example - EXCEPTION: function_call turns may include <think> tags when reasoning is added
- DO NOT generate significantly fewer or more turns than the example trajectory
- DO NOT invent or create new tools - use ONLY the provided tools

## Requirements:
1. Generate a completely NEW scenario/task that is different from the example but requires similar problem-solving patterns
2. Create a multi-turn conversation between Human and Assistant that demonstrates systematic problem-solving
3. The conversation should show the agent's reasoning process and step-by-step approach
4. **Start directly with a HUMAN message - do not include the SYSTEM content**

## Output Format:
Generate the conversation:
HUMAN: [user message content]
ASSISTANT: [assistant reply content]
HUMAN: [user message content]
ASSISTANT: [assistant reply content]
HUMAN: [user message content]
ASSISTANT: [assistant reply content]
...(until the task is finished and the conversation is complete)
\end{tcblisting}
\end{tcolorbox}
\caption{Prompt to synthesize APIGen-MT and Agenttunig}
\label{fig:Prompt to synthesize APIGen-MT}
\end{figure*}

\begin{figure*}[htbp]
    \centering
\begin{tcolorbox}[title=Prompt to synthesize OfficeBench]
\begin{tcblisting}{
  title=,
  listing only,
  breakable,
  enhanced,
  colback=white,
  colframe=black!15,
  boxrule=0.3pt,
  listing options={
    basicstyle=\ttfamily\scriptsize, % ← 比 footnotesize 更小
    breaklines=true,
    columns=fullflexible
  }
}
You are an AI assistant that generates multi-turn conversation data. Based on the provided example conversation, create a new conversation that expands the task to use multiple apps in a natural workflow.

Example conversation:\{seed\_sample\}

Your task:
1. Analyze the example task
2. Create a new task that naturally extends this workflow to use multiple different apps
3. The new task should be realistic and make sense - don't force apps together artificially
4. New created task should be diverse 

Available Tools:
{availabel tools}

## CRITICAL REQUIREMENTS - AVOID COMMON ERRORS:

### Critical Error Pattern - Insufficient Path/Directory Validation:
**Root Cause: Failure to perform directory ls or mkdir validation before operations**
- **File Path/Directory Errors**: Working directory or data directory spelling, hierarchy does not match expectations, no prior checking or directory creation
- **Mandatory Operation**: Before any file operations, use ls to confirm current directory structure and file existence
- **Directory Creation**: If creating new directories, must first use mkdir -p to ensure directory exists

## Workflow Best Practices:
1. **File Discovery**: Use shell commands (ls, find) to discover actual file names before operations
2. **App Context**: Always switch\_app before using different app APIs
3. **Data Operations**: Count/filter accurately, verify each step, preserve headers
4. **File Creation**: Use proper app APIs (Excel: new_file, Word: new_document), NOT shell touch

Requirements:
1. Follow the same conversation format as the example (\#\#Task: format, <think> and <answer> structure)
2. Generate a completely new task - don't copy the example
3. Make sure the workflow feels natural and logical
4. DON'T specify exact file names in the task - let the workflow discover them with shell commands
5. Include app switching with "Successfully switched to app: [app\_name]. Available actions:" messages
6. End with finish\_task action

Create a conversation that shows the complete workflow from start to finish with proper error prevention.

Please generate the conversation content directly in this format:
HUMAN: [human message content]
GPT: [GPT reply content]
HUMAN: [human message content] 
GPT: [GPT reply content]
... 

\end{tcblisting}
\end{tcolorbox}
\caption{Prompt to synthesize OfficeBench}
\label{fig:Prompt to synthesize OfficeBench}
\end{figure*}

\section{Full Prompt for RL on the simulated environment}
\label{appendix:prompt_rl}
Please see Figure \ref{fig:Prompt to simulated OfficeBench Environment Feedback} for prompt to simulated RL environment feedback and Figure \ref{fig:Prompt to compute OfficeBench RL Reward} for the prompt to compute the reward.

\begin{figure*}[htbp]
    \centering
\begin{tcolorbox}[title=Prompt to simulated OfficeBench Environment Feedback]
\begin{tcblisting}{
  title=,
  listing only,
  breakable,
  enhanced,
  colback=white,
  colframe=black!15,
  boxrule=0.3pt,
  listing options={
    basicstyle=\ttfamily\scriptsize, % ← 比 footnotesize 更小
    breaklines=true,
    columns=fullflexible
  }
}
You are an office environment simulator that needs to generate realistic environment feedback based on eval model's actions.

Current application: {self.current_app or "None selected"}

Available applications and action formats:
{action_formats}

{response_guidance}

Testbed Data of the task:
{testbed_text}

Please simulate the execution result of this action. You need to:
1. FIRST check if the action is valid according to the available action formats above
2. If the action is invalid (wrong app, wrong action name, missing parameters), return failure immediately. 
3. If valid, simulate realistic execution results
4. Return appropriate observation results

IMPORTANT: Only actions listed in the "Available applications and action formats" section are valid. If the action is not in that list, it should fail.

Response format:
```json
{{
    "success": true/false,
    "observation": "Specific observation result explaining success or failure."
}}
Please ensure your response is realistic. Invalid actions should always return success=false.

Current Task:
{current_task}

Previous history of the eval model's actions:
{self._get_interaction_history()}

Action to simulate:
{eval_context}
"""
\end{tcblisting}
\end{tcolorbox}
\caption{Prompt to simulate OfficeBench Environment Feedback}
\label{fig:Prompt to simulated OfficeBench Environment Feedback}
\end{figure*}

\begin{figure*}[htbp]
    \centering
\begin{tcolorbox}[title=Prompt to compute OfficeBench RL Reward]
\begin{tcblisting}{
  title=,
  listing only,
  breakable,
  enhanced,
  colback=white,
  colframe=black!15,
  boxrule=0.3pt,
  listing options={
    basicstyle=\ttfamily\scriptsize, % ← 比 footnotesize 更小
    breaklines=true,
    columns=fullflexible
  }
}
You need to evaluate whether the eval model has successfully completed the given task based on the interaction history.

Complete interaction history:
{interaction_history}

Please analyze the above information and determine whether the model has successfully completed the task based on the trajectory history and whether the final outcome meets the requirements if following all the actions. Do NOT simply judge success based on whether the model claimed "task finished". 

Please provide judgment in the following format:
```json
{{
    "reasoning": "Detailed explanation of your judgment about whether the task is successfully completed",
    "evidence": "Key evidence from the interaction history that supports your judgment",
    "task_success": true/false,
    "confidence": 0.0-1.0
}}
\end{tcblisting}
\end{tcolorbox}
\caption{Prompt to compute OfficeBench RL Reward}
\label{fig:Prompt to compute OfficeBench RL Reward}
\end{figure*}

\begin{figure*}[htbp]
    \centering
\begin{tcolorbox}[title=Prompt to simulate APIGen-MT Environment Feedback]
\begin{tcblisting}{
  title=,
  listing only,
  breakable,
  enhanced,
  colback=white,
  colframe=black!15,
  boxrule=0.3pt,
  listing options={
    basicstyle=\ttfamily\scriptsize,
    breaklines=true,
    columns=fullflexible
  }
}
You are a simulation environment. Based on the RL model's response, you need to simulate the human or tool response.
System prompt (task description and rules):
{self.system_prompt}
Reference conversation examples (use this as reference data):
{ref_conv_text}
Current conversation history:
{history_text}
RL model's latest response:
{agent_message}
Requirements:
1. Based on the reference conversation and system prompt, determine how to respond
2. **Important for tool call format checking**: 
   - If the RL model is attempting to call a tool, check if it is properly wrapped in <tool_call></tool_call> tags
   - If the tool call is NOT in <tool_call> tags or has incorrect format (e.g., malformed JSON, wrong structure), return an error message like: "Error: Tool call must be wrapped in <tool_call></tool_call> tags with proper JSON format"
   - Only if the tool call format is correct, proceed to generate the tool response
3. **Important for tool responses**: 
   - If the RL model called a tool correctly, check if related results exist in the reference conversation
   - If YES: Include the related results from the reference conversation in the tool response
   - If NO: Generate reasonable results for the tool response based on the query parameters
   - If the format is incorrect, return an error message explaining why
3. If the RL model is waiting for human input, simulate the human's next message based on the reference conversation flow
4. Follow the pattern in the reference conversation examples, but adjust appropriately to fit the current conversation
5. If the task is completed or the conversation should end, output the "[TERMINATE]" marker
Output format:
- If continuing the conversation, output the message content directly
- If should end, output "[TERMINATE]"
Please directly generate the next message without the prefix "User/Tool response":
\end{tcblisting}
\end{tcolorbox}
\caption{Prompt to simulate APIGen-MT Environment Feedback}
\label{fig:prompt_simulate_environment_feedback}
\end{figure*}

\begin{figure*}[htbp]
    \centering
\begin{tcolorbox}[title=Prompt to compute APIGen-MT RL Reward]
\begin{tcblisting}{
  title=,
  listing only,
  breakable,
  enhanced,
  colback=white,
  colframe=black!15,
  boxrule=0.3pt,
  listing options={
    basicstyle=\ttfamily\scriptsize,
    breaklines=true,
    columns=fullflexible
  }
}
You need to evaluate the RL model's performance in this task and provide a reward.
System prompt (rules and policies):
{self.system_prompt}
Complete conversation history:
{history_text}
Evaluation criteria:
1. Did the RL model correctly understand the task requirements
2. Did the RL model follow the rules and procedures in the system prompt
3. **Tool call format check**: If the RL model attempted to call tools, were they properly wrapped in <tool_call></tool_call> tags with correct JSON format? If tool calls were malformed or not properly tagged, this should be considered a FAILURE.
4. Did the RL model handle tool responses correctly
**IMPORTANT**: 
- Simply completing the user's request is NOT sufficient if policies were violated
- Incorrect tool call format (missing <tool_call> tags or malformed JSON) should result in reward = 0
Output requirement:
Please output the evaluation result in JSON format:
{{
    "reasoning": "Detailed evaluation reasoning, explaining why this reward was given"
    "reward": 0 or 1,  // 0 means failure, 1 means success
}}
Please perform the evaluation:
\end{tcblisting}
\end{tcolorbox}
\caption{Prompt to compute APIGen-MT RL Reward}
\label{fig:prompt_compute_rl_reward}
\end{figure*}

\section{More Results of Synthetic Data}
\label{More Results of Synthetic Data}
\begin{figure*}
    \centering
    \includegraphics[width=0.6\textwidth]{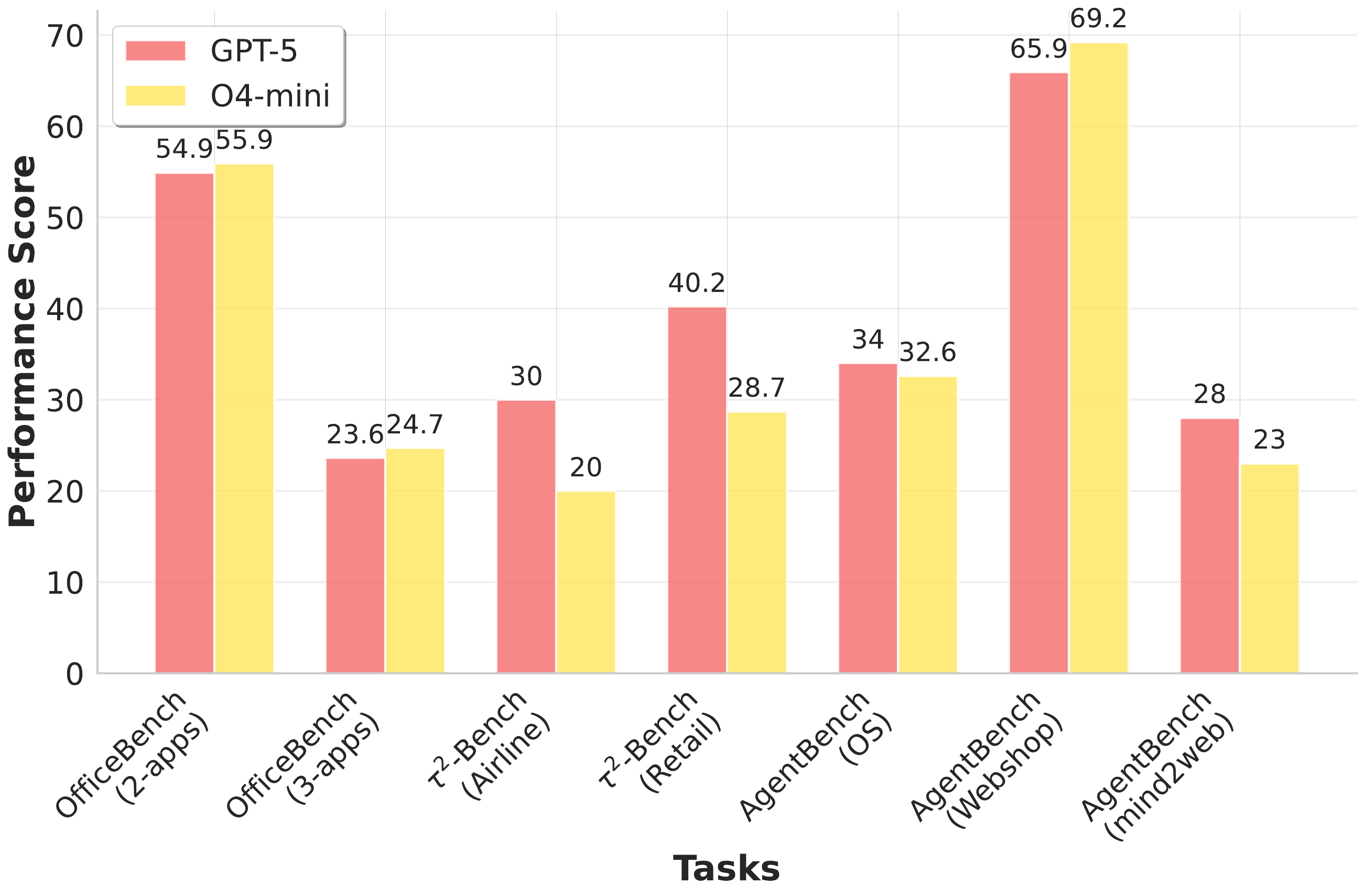}
    \caption{Ablation on different trajectory simulators. Performance comparison of 15k synthetic simulated trajectories generated by GPT-5 and o4-mini across OfficeBench, $\tau^2$-Bench, and AgentBench, showing comparable results for both of them as synthesizers across most tasks.}
    \label{fig:ablation_on_synthesizor}
\end{figure*}
We compare the performance of 15k synthetic trajectories generated by GPT-5 and o4-mini across OfficeBench, $\tau^2$-bench, and AgentBench, as shown in Figure~\ref{fig:ablation_on_synthesizor}. We fine-tune their simulated trajectories on Qwen2.5-7B-Instruct model. As the simulated trajectory synthesizer, the performance results of o4-mini and GPT-5 are similar across most tasks, with o4-mini slightly outperforming GPT-5 on OfficeBench 2-apps (55.9 vs. 54.9), 3-apps (24.7 vs. 23.6), and Webshop (69.2 vs. 65.9), while GPT-5 excels on AgentBench mind2web (28 vs. 23). Notably, GPT-5 achieves higher scores on $\tau^2$-bench, particularly in Retail (40.2 vs. 28.7) and airline (30 vs. 20) domains.

\begin{figure*}
    \centering
    \includegraphics[width=0.8\textwidth]{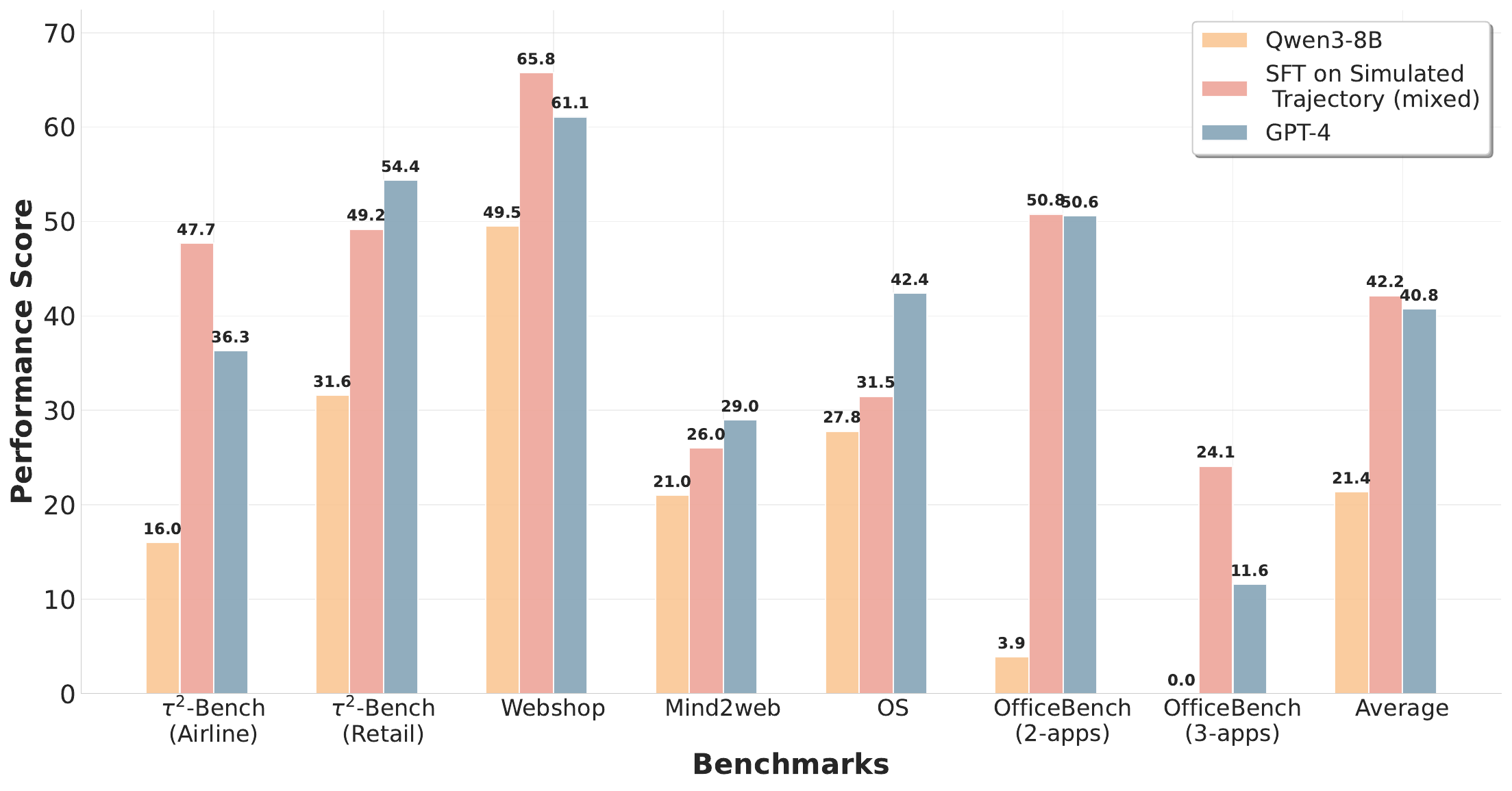}
    \caption{Performance of Qwen3-8B fine-tuned on combined three datasets generated by our pipeline jointly. Averaged across all benchmarks, our SFT models achieve higher scores than GPT-4, highlighting the effectiveness of simulation-based synthetic data for improving agent performance.}
    \label{fig:joint}
\end{figure*}

We train Qwen3-8B on all three datasets jointly and compare its performance against baselines, including Qwen3-8B and GPT-4. Fine-tuning on simulated trajectories yields substantial gains over the base model on several benchmarks, e.g., $\tau^2$-Bench and OfficeBench. On $\tau^2$-Bench, Airline improves from 16.0 to 47.7, exceeding GPT-4 (36.3). Averaged across all benchmarks, our SFT models achieve higher scores than GPT-4, highlighting the effectiveness of simulation-based synthetic data for improving agent performance.

\section{More Results of RL on Simulated Environment}
\label{More Results of RL on Simulated Environment}
\begin{figure*}[htbp]
    \centering
    \includegraphics[width=0.7\textwidth]{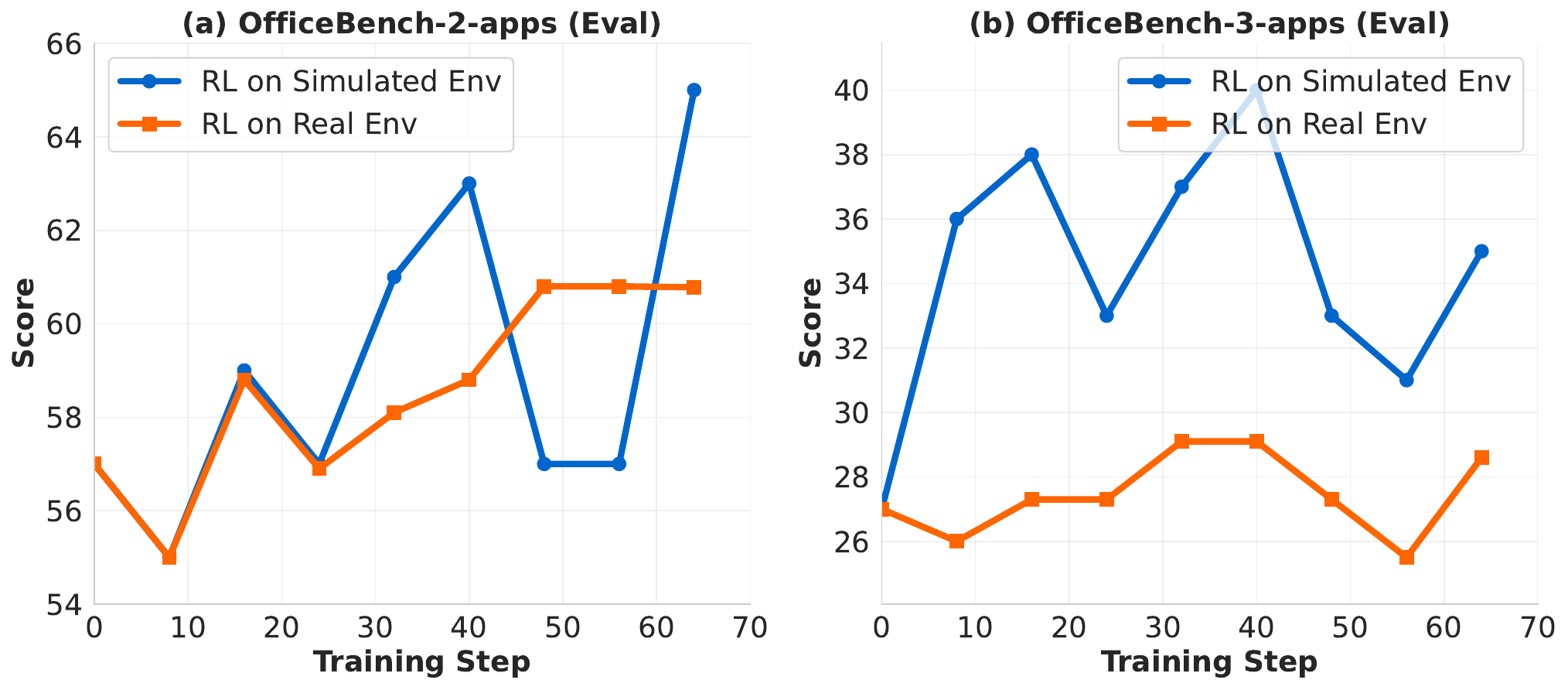}
    \caption{Performance of RL on simulated environment and real environment across different training steps for OfficeBench.}
    \label{fig:officebench_rl_training_curve}
\end{figure*}

\begin{figure*}[htbp]
    \centering
    \includegraphics[width=0.7\textwidth]{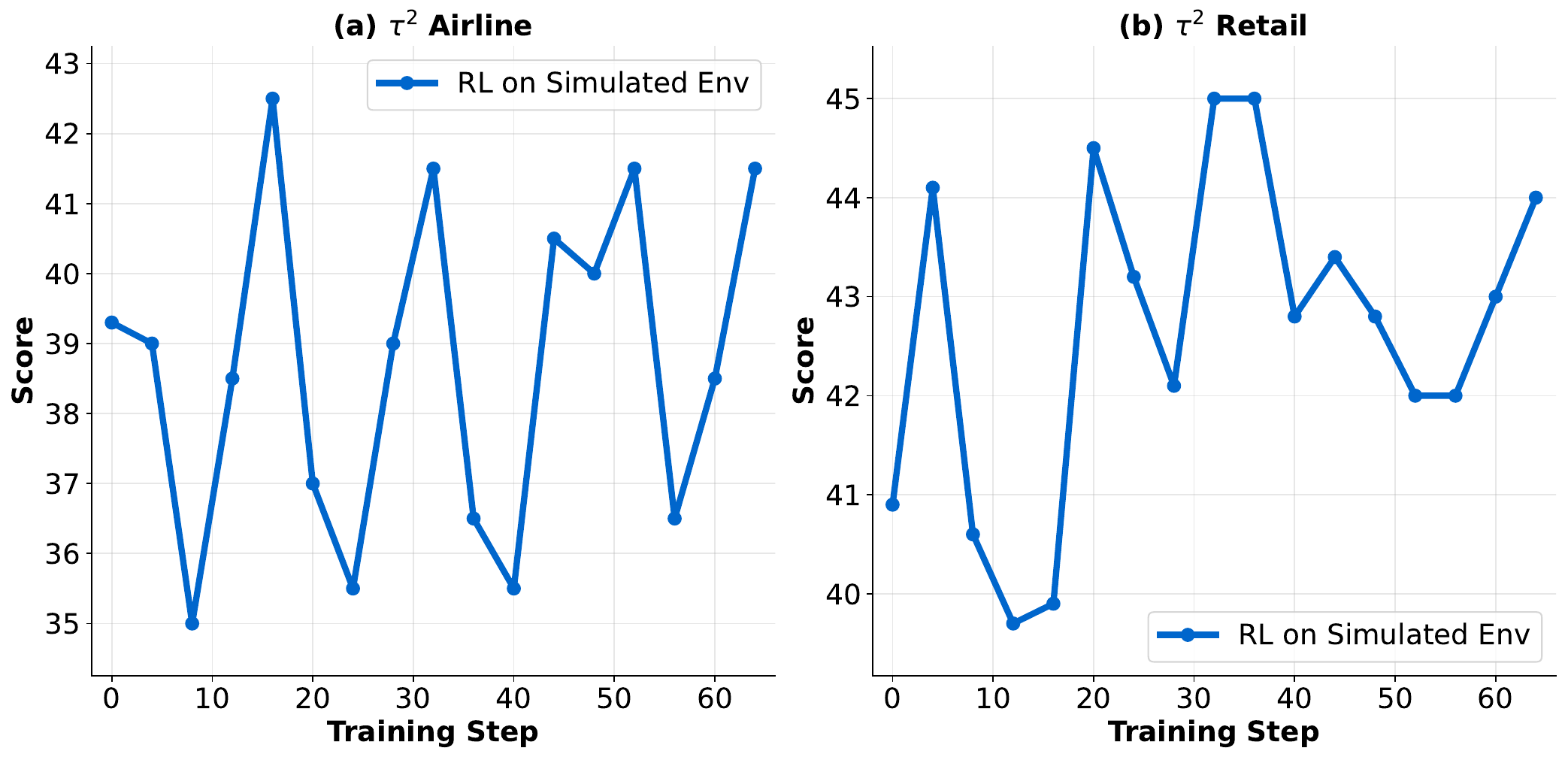}
    \caption{Performance of RL on simulated environment across different training steps for APIGen-MT.}
    \label{fig:tau2_rl_training_curve}
\end{figure*}
We present the experimental results for the RL performance on the simulated environment for OfficeBench and APIGen-MT across training steps, as depicted in Figure \ref{fig:officebench_rl_training_curve} and \ref{fig:tau2_rl_training_curve}. The evaluation performance on the OfficeBench 2-apps and 3-apps tasks demonstrates improvement, with 2-apps scores rising steadily from around 57.8 to approximately 64.7, and 3-apps scores increasing from 27.3 to 34.5. We can find that for the 3-apps performance RL on simulated environment is always better than real environment. RL on APIGen-MT also slightly improves performance on $\tau^2$-Bench.

\end{document}